%% file: main_v3.tex
\documentclass[runningheads]{llncs}
\usepackage[T1]{fontenc}
\usepackage{graphicx}
\usepackage{multirow}
\usepackage{graphicx,verbatim}
\usepackage{adjustbox}
\usepackage{multirow}
\usepackage{bm}
\usepackage[table, dvipsnames, svgnames, x11names, HTML]{xcolor}
\usepackage{tcolorbox}
\usepackage{graphicx,verbatim, xspace, amsmath, amsfonts, algorithm, algorithmicx, algpseudocode}
\usepackage{array, booktabs}
\usepackage{tikz}
\usetikzlibrary{positioning}
\usepackage{graphicx}
\usepackage{booktabs}
\usepackage{wrapfig}

\begin{document}
\title{
PhyDiCT: Plug-and-Play CT Reconstruction from Sparse X-Rays via Differentiable Rendering and Strong Priors
} 
\titlerunning{PhyDiCT: Plug-and-Play CT Reconstruction}
%
\author{Weicheng Dai\inst{1}\orcidID{0009-0009-2811-2198} \and
Shantanu Ghosh\inst{1}\orcidID{0000-0003-4085-541X} \and
Kayhan Batmanghelich\inst{1}\orcidID{0000-0001-9893-9136} \thanks{Corresponding author.}}


\authorrunning{W. Dai et al.}

\institute{
Department of Electrical and Computer Engineering,\\
Boston University, Boston, MA, USA\\
\email{\{wd2119, shawn24\}@bu.edu}\\
\email{batman@bu.edu (Corresponding author)}
}
  
\maketitle              
\newcommand{\ours}{{PhyDiCT}\xspace}
\input{sections/abstract_v2}
\input{sections/introduction_v5}

\input{sections/method_v3}

\input{sections/experiments_v2}

\input{sections/results_v3}


\input{reference.bbl}
\end{document}

%% file: sections/abstract_v2.tex
\begin{abstract}
Reconstructing 3D Computed Tomography (CT) images from a few X-ray projections is a highly ill-posed inverse problem due to the loss of volumetric information. We propose \textbf{\ours}, a
training-free framework that integrates a differentiable \textbf{Phy}sics-based forward model, grounded in the Beer–Lambert law, with a text-conditioned \textbf{Di}ffusion as a strong prior to reconstruct 3D lung \textbf{CT} images.
We refer to our approach as training-free since the prior model is used without fine-tuning, and our goal is to \emph{steer} the denoising procedure to generate samples consistent with X-ray observations. 
We guide the diffusion generation using Split Gibbs sampling to jointly optimize for projection fidelity (reward) and consistency with prior knowledge. 
Also, we introduce a test-time refinement step that enhances image realism and anatomical coherence. 
We extensively evaluate our method on publicly available 3D CT datasets using both perceptual and semantic metrics, demonstrating that it surpasses existing plug-and-play diffusion and fully trained reconstruction approaches. 
Our findings highlight that combining a strong generative prior with the underlying physics of image formation substantially improves reconstruction quality, e.g., \textbf{7.5\%} improvement on SSIM compared to full training methods.
\textit{Code will be released at \url{https://github.com/batmanlab/PhyDiCT}.}

\end{abstract}

%% file: sections/introduction_v5.tex
\section{Introduction}

Reconstructing volumetric lung CT images from a limited number of 2D X-ray projections in Cone-Beam Computed Tomography (CBCT)~\cite{Venkatesh2017ConeBC} is a highly ill-posed inverse problem. Traditional reconstruction approaches~\cite{momeni2024voxel,LACT,Qu2019SparseviewCR,10188909} are largely generic and typically require many X-ray projections. Moreover, they only weakly exploit prior knowledge about anatomy or tissue-specific structure. Learning-based methods address some of these limitations but require large collections of paired 2D–3D data. Such datasets are difficult to acquire in practice and computationally expensive to curate and train on.
In this work, we reduce the ill-posedness of the reconstruction problem by explicitly incorporating anatomy-specific priors. We leverage a pretrained generative model as a strong prior, enabling a plug-and-play reconstruction framework. In addition, a text-promptable generative model guided by radiology reports introduces semantic constraints that further regularize the solution space. Building on this idea, we propose \textbf{\ours}, which enforces 3D–2D consistency through a differentiable reward function that directly models the underlying physical projection process. This reward is used to steer the generative process at inference time.

Traditional methods for 3D reconstruction from 2D images~\cite{momeni2024voxel,LACT,Qu2019SparseviewCR} focus on regularization and total variation norms. Typically, they require an extensive number of views (\textit{e.g.},~\cite{LACT} requires $30$).
When the number of X-ray views is limited, a strong prior is required to alleviate the ill-posedness of the problem. 
Data-driven approaches implicitly learn the 3D-to-2D physical mapping from data.
A typical approach employs generative backbones such as GANs~\cite{x2ctgan,wang2024xreconlearningbasedpatientspecifichighresolution} or diffusion models~\cite{cahan2024xray2ctpagenerating3dctpa,x2ctdiff,xie2025dvgdiffusiondualviewguideddiffusion} that are fully trained on paired 3D--2D data.
However, end-to-end training makes them computationally expensive. Furthermore, the diversity of the paired training data directly influences their ability to generalize to unseen pathologies. 
Alternatively, \textbf{plug-and-play} (\textit{i.e.}, \textbf{training-free}) methods~\cite{dou2024diffusion,zheng2025ensemble,wu2024principled,chung2024decomposed,zhu2023denoising} steer the generation process of a pretrained generative model without end-to-end training, making them computationally efficient.
Moreover, they utilize generative models pretrained directly on the target modality (\textit{e.g.}, CT images) without paired observations (\textit{e.g.}, X-rays), which improves generalization since collecting diverse data within a single modality is less difficult.
They typically define a reward function that enforces consistency between generated samples and observations (i.e., X-ray views) to approximate posterior sampling. 
However, existing approaches suffer from critical limitations.
First, they often rely on diffusion models pretrained on natural images, which fail to capture the anatomical details and semantic fidelity required in medical imaging. 
Second, they usually leverage unconditional priors, which broaden the solution space and reduce fine-grained control over synthesis. 
Finally, their reward functions lack physics-based grounding, ignoring the 3D-to-2D image formation process.
These limitations motivate the use of stronger text-guided generative priors to constrain the output space and a more reliable reward model to bridge the domain gap between 2D X-rays and 3D CT images.

\ours method adopts a plug-and-play paradigm to reconstruct lung CT volumes from sparse X-ray views.
We formulate the problem as posterior sampling from a pre-trained diffusion model that is \emph{tilted} by a reward function.
Our reward function implements the Beer--Lambert Law~\cite{BBLLaw} as a differentiable X-ray renderer that simulates the 3D-to-2D projection process. 
Our method requires a generative model, and more specifically a diffusion model, that directly outputs full 3D volumes.
Since our reward function operates on full CT images, we adopt a diffusion model that directly generates full CT volumes rather than models based on latent diffusion. Models based on latent diffusion require backpropagation through a decoder, which tends to be computationally expensive in terms of GPU memory. 
For this reason, we adopt MedSyn~\cite{medsyn2024}. MedSyn also allows conditioning on radiology reports, which can further constrain the prior distribution. This is particularly useful when the radiology report associated with the X-ray is available. 
Our control mechanism is inspired by Split Gibbs Sampling~\cite{10.1109/TSP.2019.2894825}, alternating between likelihood optimization and denoising. 
We develop an \emph{Extra Compute} (EC) refinement strategy to avoid reconstruction artifacts.
In summary, our contributions are \textbf{threefold}: (1) we present a plug-and-play volumetric lung CT reconstruction method from sparse X-rays using a strong text-guided prior, (2) we integrate the physics-based 3D-to-2D imaging process into the reward function, and 
(3) we propose an EC refinement procedure to suppress artifacts. 
We rigorously evaluate \ours across perceptual, semantic, and domain-specific metrics, outperforming all baselines.

%% file: sections/method_v3.tex
\begin{figure}[t]
    \centering
\includegraphics[width=\linewidth]{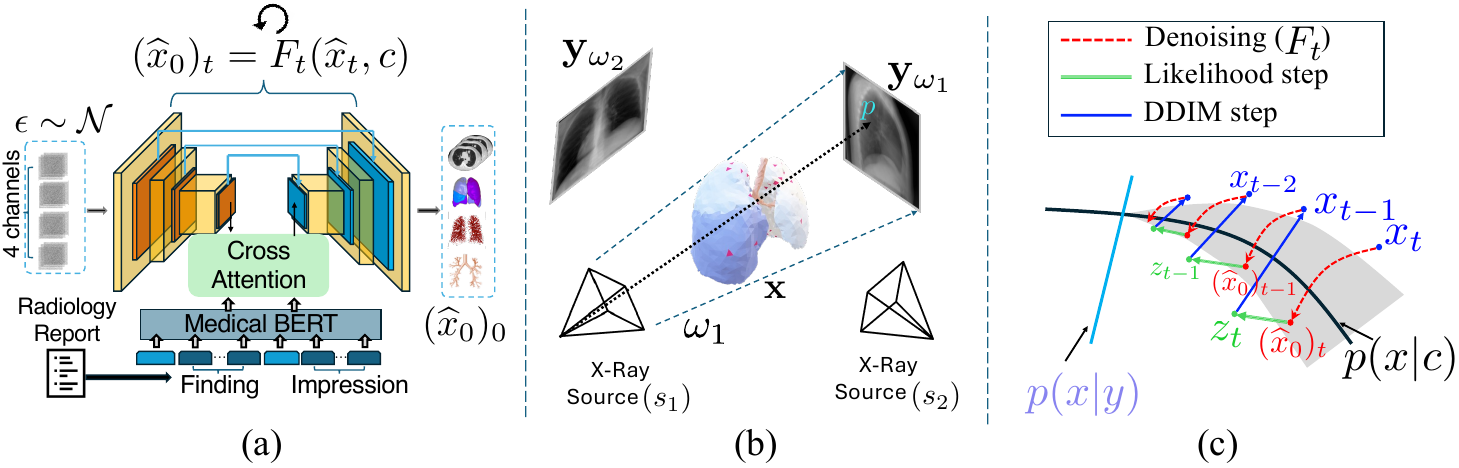}    
    \caption{
    \textbf{(a): Denoiser stage 1.}
    The text prompt $c$ (findings and impression tokens with \texttt{<CLS>} and \texttt{<SEP>} tokens) conditions the iterative generation of image $(\widehat{x}_0)_0$,
    which is later used for stage 2.
    \textbf{(b): Differentiable X-ray rendering.}
    Real-world setup where multiple X-rays (each $\omega$ starting at the same $s$, ending at multiple $p$ on detector) are used to acquire one entire $y$ on the receiver plane.
     \textbf{(c): Overview of reward-guided reconstruction.}
    The \colorbox{gray!20}{shadow} illustrates distribution envelope of $(\widehat{x}_0)_t$, which shrinks as it approaches $p(x|y)$. 
    A denoiser $F_t$ is first applied to move the noisy sample $x_t$ (lying outside the manifold) toward the clean data distribution $p(x|c)$. 
    Subsequent likelihood step shifts the predicted $(\widehat{x}_0)_t$ to a new variable $z_t$ (maybe lying outside) guided by the reward function to drive it closer to the posterior space $p(x|y)$.
    }
    \label{fig:sgs_overview}
\end{figure}

\section{Method}

\label{sec:method}
Given a set of projection angles $\boldsymbol{\omega} = \{\omega_1, \omega_2, \dots, \omega_K\}$ defining source--detector geometries, we obtain multiple 2D projections $\mathcal{Y} = \{y_{\omega_k}\}_{k=1}^K$. Our goal is to sample from the posterior $p(x|\mathcal{Y}, c) \propto p(x|c) e^{-J(\mathcal{Y},x)}$, where $p(x|c)$ is the prior distribution (\textit{i.e.,} MedSyn) with an optional text condition $c$, and $-J(\mathcal{Y},x)$ is a reward measuring consistency with the observations. 
Our method iteratively steers the denoising process of a diffusion model to tilt it toward maximizing the reward function. Finally, we perform Extra Compute (EC) refinement to remove reconstruction artifacts.

\noindent \textbf{Prior Distribution (MedSyn).}
MedSyn~\cite{medsyn2024} is a diffusion model based on the DDPM~\cite{song2021denoising}  framework that optionally accepts a text condition ($c$). Along with images, it also produces lung segmentation masks, which increase the anatomical fidelity of the generated images. Starting from pure noise $x_T$, it iteratively alternates between two steps at each timestep $t$ during inference. First, it uses a denoiser network ($F_t$) to predict the clean image $(\widehat{x}_0)_t$ based on the noisy counterpart $x_t$ and the optional text condition $c$. Given the current estimate of the clean image $(\widehat{x}_0)_t$, it then applies the deterministic DDIM backward step~\cite{song2021denoising} to compute the next sample $x_{t-1}$.
\begin{align}
(\widehat{x}_0)_t &= F_t(x_t, c), \label{eq:medsyn1} \\
    x_{t-1} = \text{DDIM}\left((\widehat{x}_0)_t, x_t\right) &= \sqrt{\bar{\alpha}_{t-1}}(\widehat{x}_0)_t 
    + \sqrt{1-\bar{\alpha}_{t-1}}
    \frac{x_t - \sqrt{\bar{\alpha}_t}(\widehat{x}_0)_t}{\sqrt{1-\bar{\alpha}_t}}, 
\label{eq:ddim}
\end{align}
where $\bar{\alpha}_t = \prod_{i=1}^{t}\alpha_i$ represents the cumulative variance schedule, and $c$ is an optional conditioning variable (\textit{i.e.}, text). Repeating Eqs.~\ref{eq:medsyn1} and~\ref{eq:ddim} from $t = T$ to $0$ produces a final high-resolution CT image.

\noindent \textbf{Reward from Differentiable X-ray Rendering.} \label{sec:drr}
We leverage the fact that MedSyn predicts the clean image at each step $(\widehat{x}_0)_t$. Our goal is to design a reward function that steers the current estimate toward consistency with the observations $\mathcal{Y}$. To this end, we construct a physics-based reward function. Following~\cite{gopalakrishnan2022fast}, we model the X-ray formation process using the Beer--Lambert law. Let the ray $\mathbf{r}(\lambda) = \mathbf{s} + \lambda(\mathbf{p} - \mathbf{s})$ denote a photon path from the X-ray source $\mathbf{s} \in \mathbb{R}^3$ to a pixel $\mathbf{p} \in \mathbb{R}^3$ on the detector plane, where $\lambda \in [0,1]$. 
The observed intensity at $\mathbf{p}$ is proportional to the accumulated linear attenuation coefficient $\mu(\cdot) : \mathbb{R}^3 \rightarrow [0, \infty)$ along the spatial path through the volume. In practice, CT images are represented as discrete voxel grids $x \in \mathbb{R}^{H \times W \times D}$ that approximate $\mu(\cdot)$; therefore, we discretize the integral as a summation (with $\mathbf{s}$ omitted):
\begin{align}
I_{\mu}(\mathbf{r}) 
&= \|\mathbf{p} - \mathbf{s}\| 
\int_0^1 \mu(\mathbf{s} + \lambda(\mathbf{p} - \mathbf{s})) \, d\lambda,
\label{eq:beer} \\[0.5em]
\mathcal{P}(x)[\mathbf{p}] 
&= \|\mathbf{p} - \mathbf{s}\| 
\sum_{m=1}^{M-1} 
x\!\left(\mathbf{s} + \tfrac{\lambda_{m+1} + \lambda_m}{2}(\mathbf{p} - \mathbf{s})\right)
(\lambda_{m+1} - \lambda_m),
\label{eq:drr}
\end{align}
where $\{\lambda_m\}_{m=1}^M$ denote the coordinates of the intersection points of the ray $\mathbf{r}(\lambda)$ with the parallel voxel planes of the 3D grid. 
$\mathcal{P}: \mathbb{R}^{H \times W \times D} \rightarrow \mathbb{R}^{H' \times W'}$ defines a differentiable projection operator that maps the 3D attenuation field to a 2D X-ray image via trilinear interpolation.

Since Eq.~\ref{eq:drr} approximates the actual X-ray formation process, a domain shift may exist between the observations and $\mathcal{P}(x)$. To mitigate this effect, we incorporate the LPIPS loss~\cite{zhang2018perceptual} as an additional regularization of guidance to further enforce latent feature consistency. For an observation $y_{\omega_k} \in \mathcal{Y}$ acquired at angle $\omega_k \in \boldsymbol{\omega}$, the total loss (i.e., the negative reward) is defined as the sum of the reconstruction and LPIPS losses:
\begin{equation}
J(\mathcal{Y},x') = \frac{1}{K} \sum_{k} \left( ||y_{\omega_k}- \mathcal{P}_{\omega_k}(x')||_2^2 +  \text{LPIPS}(\mathcal{P}_{\omega_k}(x'), y_{\omega_k}) \right).
\label{eq:proj}
\end{equation}

\label{sec:method_SGS}
\noindent \textbf{Reward-guided Reconstruction.}
We treat the reconstruction as sampling from the posterior distribution $p(x|\mathcal{Y},c)$, which can be viewed as a prior distribution $p(x|c)$ tilted by the reward $e^{-J(\mathcal{Y},x)}$, i.e., $p(x|\mathcal{Y},c) \propto p(x|c)\, e^{-J(\mathcal{Y},x)}$.
To sample from this posterior distribution, we adopt Split Gibbs Sampling (SGS)~\cite{10.1109/TSP.2019.2894825}, which introduces an auxiliary random variable $z$ and splits the sampling process between two distributions: $\pi(x|z, c)$ and $\pi(z|x, \mathcal{Y})$:

\begin{subequations}\label{eq:sgs}
\noindent
\begin{minipage}{0.45\linewidth}
\begin{equation}
\underbrace{
\pi(x|z, c) \propto 
\exp\left(-g(x,c) - \frac{1}{\rho}\|x-z\|_2^2\right)}_{\text{Denoising}}, \notag
\end{equation}
\end{minipage}
\hfill
\begin{minipage}{0.45\linewidth}
\begin{equation}
\underbrace{\pi(z|x, \mathcal{Y}) \propto 
\exp\left(-J(\mathcal{Y},z) - \frac{1}{\rho}\|z-x\|_2^2\right)}_{\text{Likelihood}}, \notag
\end{equation}
\end{minipage}
\end{subequations}

\noindent where $g(x,c)$ as the unnormalized negative log-likelihood of the prior (\textit{i.e.}, $p(x|c) \propto e^{-g(x,c)}$) and $\rho$ a positive hyperparameter controlling the dissimilarity between $z$ and $x$. 


Sampling from $\pi(x|z, c)$ can be viewed as a denoising step in MedSyn, where $z$ is a noisy observation of $x$. In step $t$, $z_t$ is first used in Eq.~\ref{eq:ddim} to obtain $x_{t-1}$, \textit{i.e.,} $x_{t-1}=\text{DDIM}(z_t, x_{t})$. 
We then use the denoiser $F_t$ to compute the prior MAP estimate of the clean image prediction according to Eq.~\ref{eq:medsyn1}.
The resulting estimate $(\widehat{x}_0)_{t-1}$ serves as $x$ for the subsequent likelihood step.

Sampling from $\pi(z|x, \mathcal{Y})$ can be viewed as a likelihood step that encourages higher fidelity to the observations while remaining close to $x$. 
We approximate this step by finding an optimal $z$ that minimizes the negative log-likelihood:
$\ell(z) = J(\mathcal{Y},z) + \frac{1}{\rho}\|z-(\widehat{x}_0)_{t-1}\|_2^2.$
Instead of performing a full and costly sampling procedure, we take a few gradient steps (two in all experiments) on $\ell(z)$ to obtain an optimized $z_{t-1}$. This solution balances fidelity to the observations $\mathcal{Y}$ and consistency with the prior prediction $(\widehat{x}_0)_{t-1}$. The entire procedure is visualized in Fig.~\ref{fig:sgs_overview}.

\noindent\textbf{Extra Compute (EC) Refinement.}
\label{sec:EC}
We empirically observed that the final guided output $(\widehat{x}_0)_0$ shows minor artifacts (Fig.~\ref{fig:inverse3d_ct_imgs}). 
This suggests the resulting image, while faithful to the observations, may lie off the data manifold learned by the prior. 
We introduce an additional step to ensure the final output lies on this manifold. Selecting a noise level according to timestep $\delta T$, where $\delta \in (0,1)$, 
we first apply the forward diffusion process to $(\widehat{x}_0)_0$ to obtain a noisy image:
    $x_{\delta T} = \sqrt{\bar{\alpha}_{\delta T}}(\widehat{x}_0)_0 + \sqrt{1-\bar{\alpha}_{\delta T}}\epsilon \label{eq:diffusion_forward},$
where $\epsilon \sim \mathcal{N}(0, I)$. 
We then apply the standard backward denoising process (Eqs.~\ref{eq:medsyn1}-\ref{eq:ddim}) from $t=\delta T$ to $t=0$ \textit{without} the likelihood guidance step. 
This step effectively projects the image back onto the prior's data manifold, resulting in better quality.

%% file: sections/experiments_v2.tex
\begin{figure*}[t]
\centering
\includegraphics[width=1.0\textwidth]{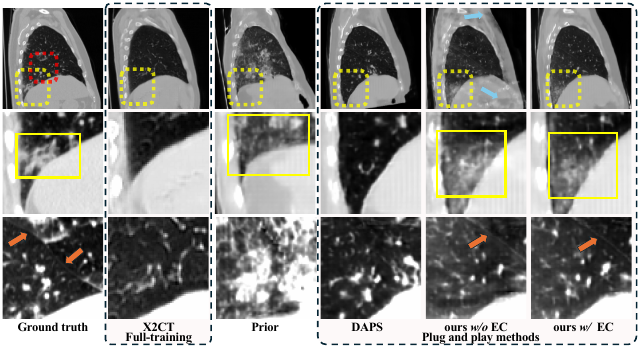}
\caption{
\textbf{Comparison of reconstructed CT based on DiffVox.}
The first row shows raw slices, and following rows are zoomed-in regions.
\textcolor{blue!30!white}{Arrows} on first row show minor artifacts, further suppressed with EC.
Second row (\textcolor{yellow!80!black}{bounding box}) highlights that our method accurately reconstructs subtle pathology, while avoiding overemphasis of disease compared to Prior.
Third row (\textcolor{red}{arrows}) highlights the anatomical reconstruction of fissure lines.
}
\label{fig:inverse3d_ct_imgs}
\end{figure*}

\section{Experiments}
We evaluate the fidelity of the generated CT images both qualitatively and quantitatively against the ground truth, for which we report SSIM and PSNR. 
We also assess perceptual realism and report FID for this purpose. 
In addition, we evaluate how closely the generated images are consistent with pathology by reporting CLIP-based semantic scores~\cite{hamamci2024foundation} using both text-to-image (T2I) and image-to-image (I2I) similarity. We compare against \textit{(1) full-training} methods specifically designed for this problem, namely Xray2CTPA~\cite{cahan2024xray2ctpagenerating3dctpa} and X2CT~\cite{x2ctgan}, and \textit{(2) plug-and-play baselines}, including PnP-DM~\cite{wu2024principled} and DAPS~\cite{ZhangCBMA025}. For the plug-and-play methods, we follow the exact same setup as ours to ensure a fair comparison. We also perform ablation studies with respect to the number of views and the effect of the EC component.

\noindent \textbf{Dataset Preparation and Experimental Setup.}
We sample 400 3D CT images paired with radiology reports from CT-Rate~\cite{hamamci2024foundation} as our evaluation dataset. 
Since there is no public dataset with paired CT and X-ray images, we synthesize X-ray images (following~\cite{10635648,x2ctgan}) using two methods: DiffVox~\cite{momeni2024voxel} and DeepDRR~\cite{unberath2018deepdrr}. While DiffVox is more consistent with our reward function, DeepDRR simulates more realistic scenarios, including noise and beam-hardening effects. The projected X-ray images in all experiments have a size of $448 \times 448$, with projection views horizontally spanning $\left\{ \frac{k}{K}2\pi \right\}_{k=0}^{K-1}$ for $K$ views (default $K=8$), from a fixed distance of $1020\,\mathrm{mm}$ between the X-ray source and the receiver plane, following~\cite{momeni2024voxel}. 
Ideally, the text condition should be derived from X-ray reports. 
Due to the lack of such public datasets, we use radiology reports from CT images as proxies for X-ray reports, as most major pathologies are observable in both CT and X-ray imaging.
We also conduct experiments without radiology reports.
We follow the MedSyn preprocessing pipeline, including image alignment and intensity normalization (MinMax norm into $[-1, 1]$).
Prior model (MedSyn) generates low resolution images of size $64^3$ with text $c$, followed by high resolution of $256^3$.
\ours is applied on both branches to ensure the knowledge from both $c$ and $y$ (trilinear interpolation $4\times$ on low resolution when computing the reward). 
We adopt $\rho=1.0$ for all experiments, $2$ optimizing steps during each iteration, and learning rate $0.1$ following~\cite{wu2024principled,ZhangCBMA025}. 
(Hardware: single NVIDIA RTX 6000 GPU; Software: Pytorch 2.6.0+cu126).

%% file: sections/results_v3.tex

\definecolor{RowBlue}{HTML}{E8EBFF}
\definecolor{RowGreen}{HTML}{EAF6EA}

\definecolor{RowBlue}{HTML}{E8EBFF}
\definecolor{RowGreen}{HTML}{EAF6EA}

\begin{table*}[t]
\caption{\textbf{Quantitative results.}
The 1st and 2nd best are marked in \textbf{bold} and \underline{underline} (excluding Ground Truth). 
\textcolor[HTML]{A0A8FF}{Blue} lines denote full training model, while \textcolor[HTML]{7FC47F}{Green} lines denote plug-and-play models.
\ours prevails over the other baselines, except on par with \textcolor[HTML]{A0A8FF}{Blue} baselines in Semantic aspects. 
}
\begin{adjustbox}{max width=\textwidth, center}
\begin{tabular}{clcccccccc}
\toprule

\multicolumn{2}{c}{} &
\multicolumn{6}{c}{\textbf{Perceptual}} &
\multicolumn{2}{c}{\textbf{Semantic}}  \\

\cmidrule(lr){3-8}
\cmidrule(lr){9-10}

\multicolumn{1}{c}{} &
\multicolumn{1}{c}{\textbf{Sources}} &
\textbf{SSIM}$\uparrow$ &
\textbf{PSNR}$\uparrow$ &
\textbf{FID-XY}$\downarrow$ &
\textbf{FID-YZ}$\downarrow$ &
\textbf{FID-ZX}$\downarrow$ &
\textbf{FID(avg)}$\downarrow$ &
\textbf{Clip-T2I(\%)}$\uparrow$ &
\textbf{Clip-I2I(\%)}$\uparrow$ \\ 
\midrule

& Ground Truth 
& - & - 
& - & - & - & -
& $16.46$ & - \\

& Prior~\cite{medsyn2024}
& $0.2466$ & $12.29$ 
& ${4.61}$ & ${3.80}$ & $7.14$ & ${5.18}$
& $3.40$ & $78.50$ \\

\midrule










\multirow{7}{*}{\rotatebox[origin=c]{90}{\parbox{2cm}{\centering \textit{w/}$y$ from DiffVox~\cite{momeni2024voxel}}}}
&
\cellcolor{RowBlue} Xray2CTPA~\cite{cahan2024xray2ctpagenerating3dctpa}
& \cellcolor{RowBlue} $0.3421$
& \cellcolor{RowBlue} $12.54$
& \cellcolor{RowBlue} $\underline{4.71}$
& \cellcolor{RowBlue} $\underline{5.07}$
& \cellcolor{RowBlue} $7.21$
& \cellcolor{RowBlue} $\underline{5.66}$
& \cellcolor{RowBlue} $\underline{13.85}$
& \cellcolor{RowBlue} $81.19$
\\

&
\cellcolor{RowBlue} X2CT~\cite{x2ctgan}
& \cellcolor{RowBlue} $0.3969$
& \cellcolor{RowBlue} $14.65$
& \cellcolor{RowBlue} $8.61$
& \cellcolor{RowBlue} $7.52$
& \cellcolor{RowBlue} $10.14$
& \cellcolor{RowBlue} $8.76$
& \cellcolor{RowBlue} $\mathbf{16.16}$
& \cellcolor{RowBlue} $\mathbf{89.04}$
\\

&
\cellcolor{RowGreen} DAPS~\cite{dou2024diffusion}
& \cellcolor{RowGreen} $0.3641$
& \cellcolor{RowGreen} $12.76$
& \cellcolor{RowGreen} $6.61$
& \cellcolor{RowGreen} $7.93$
& \cellcolor{RowGreen} $8.07$
& \cellcolor{RowGreen} $7.54$
& \cellcolor{RowGreen} $13.25$
& \cellcolor{RowGreen} $83.16$
\\

&
\cellcolor{RowGreen} PnP-DM~\cite{wu2024principled}
& \cellcolor{RowGreen} $0.2998$
& \cellcolor{RowGreen} $14.23$
& \cellcolor{RowGreen} $7.74$
& \cellcolor{RowGreen} $10.98$
& \cellcolor{RowGreen} $8.92$
& \cellcolor{RowGreen} $9.21$
& \cellcolor{RowGreen} $13.45$
& \cellcolor{RowGreen} $80.41$
\\

&
\cellcolor{RowGreen} Ours \textit{w/o} text
& \cellcolor{RowGreen} $0.3747$
& \cellcolor{RowGreen} $\underline{15.29}$
& \cellcolor{RowGreen} $6.13$
& \cellcolor{RowGreen} $7.82$
& \cellcolor{RowGreen} $9.03$
& \cellcolor{RowGreen} $7.66$
& \cellcolor{RowGreen} $11.35$
& \cellcolor{RowGreen} $\underline{85.79}$
\\

&
\cellcolor{RowGreen} Ours \textit{w/o} EC
& \cellcolor{RowGreen} $\underline{0.3983}$
& \cellcolor{RowGreen} $\mathbf{15.85}$
& \cellcolor{RowGreen} $6.79$
& \cellcolor{RowGreen} $7.71$
& \cellcolor{RowGreen} $\underline{6.92}$
& \cellcolor{RowGreen} $7.14$
& \cellcolor{RowGreen} $12.94$
& \cellcolor{RowGreen} $81.21$
\\

&
\cellcolor{RowGreen} Ours
& \cellcolor{RowGreen} $\mathbf{0.4268}$
& \cellcolor{RowGreen} $15.20$
& \cellcolor{RowGreen} $\mathbf{3.52}$
& \cellcolor{RowGreen} $\mathbf{3.88}$
& \cellcolor{RowGreen} $\mathbf{5.76}$
& \cellcolor{RowGreen} $\mathbf{4.38}$
& \cellcolor{RowGreen} $13.75$
& \cellcolor{RowGreen} $84.66$
\\








\midrule[\heavyrulewidth]

\multirow{6}{*}{\rotatebox[origin=c]{90}{\parbox{2.5cm}{\centering \textit{w/}$y$ from DeepDRR~\cite{unberath2018deepdrr}}}}
&
\cellcolor{RowBlue} Xray2CTPA~\cite{cahan2024xray2ctpagenerating3dctpa}
& \cellcolor{RowBlue} $0.3131$
& \cellcolor{RowBlue} $12.20$
& \cellcolor{RowBlue} $\underline{4.93}$
& \cellcolor{RowBlue} $6.00$
& \cellcolor{RowBlue} $\underline{6.05}$
& \cellcolor{RowBlue} $\underline{5.66}$
& \cellcolor{RowBlue} $\mathbf{15.14}$
& \cellcolor{RowBlue} $80.68$
\\

&
\cellcolor{RowBlue} X2CT~\cite{x2ctgan}
& \cellcolor{RowBlue} $0.2619$
& \cellcolor{RowBlue} $10.70$
& \cellcolor{RowBlue} $18.60$
& \cellcolor{RowBlue} $15.92$
& \cellcolor{RowBlue} $16.31$
& \cellcolor{RowBlue} $16.94$
& \cellcolor{RowBlue} $\underline{14.46}$
& \cellcolor{RowBlue} $71.98$
\\

&
\cellcolor{RowGreen} DAPS~\cite{dou2024diffusion}
& \cellcolor{RowGreen} $0.2560$
& \cellcolor{RowGreen} $\underline{12.38}$
& \cellcolor{RowGreen} $8.51$
& \cellcolor{RowGreen} $7.14$
& \cellcolor{RowGreen} $7.07$
& \cellcolor{RowGreen} $7.57$
& \cellcolor{RowGreen} $10.37$
& \cellcolor{RowGreen} $81.13$
\\

&
\cellcolor{RowGreen} PnP-DM~\cite{wu2024principled}
& \cellcolor{RowGreen} $0.2265$
& \cellcolor{RowGreen} $12.02$
& \cellcolor{RowGreen} $14.20$
& \cellcolor{RowGreen} $11.03$
& \cellcolor{RowGreen} $15.30$
& \cellcolor{RowGreen} $13.51$
& \cellcolor{RowGreen} $12.16$
& \cellcolor{RowGreen} $77.50$
\\

&
\cellcolor{RowGreen} Ours \textit{w/o} text
& \cellcolor{RowGreen} $\underline{0.3282}$
& \cellcolor{RowGreen} $12.30$
& \cellcolor{RowGreen} $5.86$
& \cellcolor{RowGreen} $\underline{4.41}$
& \cellcolor{RowGreen} $7.26$
& \cellcolor{RowGreen} $5.84$
& \cellcolor{RowGreen} $10.44$
& \cellcolor{RowGreen} $\mathbf{85.05}$
\\


&
\cellcolor{RowGreen} Ours
& \cellcolor{RowGreen} $\mathbf{0.3534}$
& \cellcolor{RowGreen} $\mathbf{12.83}$
& \cellcolor{RowGreen} $\mathbf{3.87}$
& \cellcolor{RowGreen} $\mathbf{3.39}$
& \cellcolor{RowGreen} $\mathbf{5.58}$
& \cellcolor{RowGreen} $\mathbf{4.28}$
& \cellcolor{RowGreen} $13.33$
& \cellcolor{RowGreen} $\underline{82.29}$
\\

\bottomrule
\end{tabular}
\end{adjustbox}
\label{tab:main_ct}
\end{table*}

\noindent \textbf{Quantitative Comparison.}
Table~\ref{tab:main_ct} reports the quantitative results.
Overall, \ours outperforms all baselines in perceptual quality while preserving strong semantic features.
A significant $\mathbf{7.5}\%$ improvement is achieved in SSIM compared to other baselines (ours: $\mathbf{0.4268}$ \textit{vs.} X2CT: $\mathbf{0.3969}$) without any additional training.
Among the fully trained baselines, X2CT performs slightly better than \ours in semantic metrics but remains inferior in perceptual quality.
This is because it benefits from in-domain training on the CT-Rate dataset, whereas our prior model (MedSyn) is trained on external datasets, leading to an out-of-domain effect.
However, our method consistently outperforms Xray2CTPA in SSIM and PSNR, as we explicitly model the relationship between 2D and 3D images.
\ours also consistently outperforms the plug-and-play baselines because they apply guidance on $x_t$, whereas we rely on $x_0$ prediction and directly optimize it, resulting in more effective steering.
The EC component further boosts quantitative performance.
The bottom block of Table~\ref{tab:main_ct} shows results when DeepDRR is used to generate X-ray views.
All methods exhibit a substantial drop in perceptual metrics and CLIP-I2I scores, while CLIP-T2I behaves similarly to the main results.
X2CT particularly suffers in CLIP-I2I due to GAN collapse under domain shift, whereas Xray2CTPA remains more stable due to its incorporation of CLIP features.
However, compared to other baselines, our FID and CLIP-I2I scores remain stable relative to the original source, indicating that higher-level CT features are preserved during reconstruction.
We attribute this improvement to the LPIPS loss, which enforces consistency in a high-dimensional feature space.
These results demonstrate the robustness of \ours.

\noindent \textbf{Qualitative Comparison.}
Fig.~\ref{fig:inverse3d_ct_imgs} demonstrates that \ours reconstructs CT images with finer structural details and improved preservation of pathological features.
In addition, \ours mitigates the exaggeration artifacts exhibited by the prior model, both across the entire slice (first row) and within localized regions (second row).
In comparison, other baselines fail to reconstruct the atelectasis region, as they are unable to effectively steer the predicted $x_0$.
In the third row, our method successfully reconstructs anatomical fissure lines, whereas the other methods fail to capture these structures.
Moreover, EC notably improves the visual quality of CT reconstructions (first row), further enhancing the overall performance of \ours.

\noindent \textbf{Ablation Study.} 
We study the effects of EC time $\delta$, the number of views $K$, and text guidance $c$. 
Results in Table~\ref{tab:main_ct} without EC also show consistent performance, outperforming most baselines.
Table~\ref{tab:ssim_clip_main} shows the effect of $\delta$ on EC, from which we conclude that (1) EC requires negligible time ($<30$s) compared to the main method ($1342$s), and (2) larger $\delta$ values are generally preferred. Increasing $\delta$ up to $0.9$ improves SSIM but degrades CLIP-T2I.
We select a balanced setting of $\delta=0.8$ between SSIM and CLIP-T2I, as it preserves both perceptual and semantic features.
Table~\ref{tab:ssim_clip_nview_simple} shows the effect of $K$ on our main method, where increasing $K$ improves both SSIM and CLIP-T2I.
However, the computational cost becomes non-negligible for large values (\textit{e.g.}, $16$).
We choose a balanced setting of $K=8$, as improvements become marginal beyond this point.
The experiment in Table~\ref{tab:main_ct} \textbf{without report} shows a minor degradation in SSIM and FID due to the lack of textual guidance, yet still outperforms most baselines.
The absence of textual guidance leads to an increase in CLIP-I2I but a decrease in CLIP-T2I, highlighting the trade-off between image fidelity and semantic alignment and further emphasizing the benefit of radiological prompts.

\begin{table*}[t]
\centering

\begin{minipage}{0.48\textwidth}
\centering
\caption{
\textbf{Ablation on different $\mathbf{\delta}$ in EC.}
A clear trade-off lies between the SSIM and Clip-T2I.
}
\renewcommand{\arraystretch}{1.1}
\setlength{\tabcolsep}{3.5pt}
\scriptsize
\begin{tabular}{c | c c c}
\hline
$\mathbf{\delta}$ & \textbf{SSIM}$\uparrow$ & \textbf{Clip-T2I (\%)}$\uparrow$ & \textbf{Time (s)}$\downarrow$ \\
\hline
$0.1$  & $0.3950$ & $11.74$ & $\mathbf{3}+1342$ \\
$0.2$  & $0.4005$ & $11.93$ & $\underline{5}+1342$ \\
$0.5$  & $0.4101$ & $12.32$ & $13+1342$ \\
$0.8$  & $0.4268$ & $\mathbf{13.75}$ & $20+1342$ \\
$0.85$ & $0.4281$ & $\underline{12.61}$ & $21+1342$ \\
$0.9$  & $\mathbf{0.4358}$ & $12.07$ & $22+1342$ \\
$0.95$ & $\underline{0.4287}$ & $11.67$ & $23+1342$ \\
$0.99$ & $0.3769$ & $11.69$ & $24+1342$ \\
\hline
\end{tabular}
\label{tab:ssim_clip_main}
\end{minipage}
\hfill
\begin{minipage}{0.48\textwidth}
\centering
\caption{
\textbf{Ablation on number of views $K$ without EC.}
Smaller $K$ achieves better efficiency compared to results after $K=8$.
}
\renewcommand{\arraystretch}{1.1}
\setlength{\tabcolsep}{4pt}
\scriptsize
\begin{tabular}{c | c c c}
\hline
$K$ & \textbf{SSIM}$\uparrow$ & \textbf{Clip-T2I}(\%)$\uparrow$ & \textbf{Time (s)}$\downarrow$ \\
\hline
$2$  & $0.3340$ & $8.29$  & $\mathbf{453}$ \\
$4$  & $0.3628$ & $10.99$ & $\underline{857}$ \\
$8$  & $0.3983$ & $12.94$ & $1342$ \\
$16$ & $\underline{0.4032}$ & $\underline{13.15}$ & $2740$ \\
$32$ & $\mathbf{0.4073}$ & $\mathbf{13.90}$ & $4709$ \\
\hline
\end{tabular}
\label{tab:ssim_clip_nview_simple}
\end{minipage}
\end{table*}


\section{Conclusion}
We propose \ours, reconstructing 3D lung CT images from limited view X-ray images. 
It benefits from a physical 3D-2D rendering method and a strong text-guided prior model; meanwhile involving $x_0$ prediction and LPIPS loss yields better results.
On the other hand, \ours is specifically designed for 3D lung CT reconstruction, thus for other organs we need corresponding pretrained priors.


\subsubsection*{Ethical Statement and Data Usage}
The retrospective data used in this study was obtained from the publicly available CT-Rate repository \cite{hamamci2024foundation}. All data utilization was performed in compliance with the repository's data use agreement and institutional ethical guidelines.

\subsubsection*{Acknowledgements}
This work was supported by NIH grant 5R01HL141813 and NSF CAREER grant 2443167.

\subsubsection*{Disclosure of Interests}
The authors declare no competing interests in the paper.
